\documentclass[letterpaper, 10 pt, conference]{ieeeconf}  

\IEEEoverridecommandlockouts                              

\usepackage{color, soul}
\usepackage{amsmath, amssymb}
\usepackage{graphicx}
\usepackage{siunitx}
\usepackage{booktabs}
\usepackage{multirow}
\usepackage{rotating}
\usepackage{subfig}
\usepackage{threeparttable}
\usepackage{textcomp}
\usepackage{gensymb}
\usepackage{pifont} 
\usepackage{float}

\usepackage{enumitem}
\setlist[itemize]{noitemsep,nolistsep,leftmargin=17pt}
\setlist[enumerate]{noitemsep,nolistsep,leftmargin=17pt}
\usepackage[font={small}]{caption}
\usepackage[free-standing-units=true]{siunitx}
\usepackage{xfrac}
\usepackage{url}
\usepackage{stfloats}
\usepackage[colorlinks=true,allcolors=blue]{hyperref}
\usepackage{subfig}
\usepackage[table,dvipsnames]{xcolor}

\usepackage[T1]{fontenc}
\usepackage{aecompl}
\usepackage{array}
\newcommand{\PreserveBackslash}[1]{\let\temp=\\#1\let\\=\temp}
\newcolumntype{C}[1]{>{\PreserveBackslash\centering}p{#1}}
\newcolumntype{R}[1]{>{\PreserveBackslash\raggedleft}p{#1}}
\newcolumntype{L}[1]{>{\PreserveBackslash\raggedright}p{#1}}

\newcommand{\equref}[1]{Eq.~\ref{#1}}
\newcommand{\fref}[1]{Fig.~\ref{#1}}
\newcommand{\sref}[1]{Section~\ref{#1}}
\newcommand{\tref}[1]{Table~\ref{#1}}

\newcommand{\etal}{\textit{et al.}~}
\newcommand{\ie}{{i.e.},~}

\renewcommand{\paragraph}[1]{\noindent\textbf{#1}~}

\title{\LARGE \bf
Enhancing Scene Coordinate Regression with Efficient Keypoint Detection and Sequential Information
}
\author{Kuan~Xu$^{1\star}$,~
        Zeyu~Jiang$^{1\star}$,~
        Haozhi~Cao$^{1}$,~
        Shenghai~Yuan$^{1}$,~
        Chen~Wang$^{2}$, 
        Lihua~Xie$^{1}$,~\IEEEmembership{Fellow,~IEEE}
\thanks{{$^\star$}Equal contribution.}
\thanks{This work is supported by the National Research Foundation of Singapore under its Medium-Sized Center for Advanced Robotics Technology Innovation.}
\thanks{$^{1}$Kuan Xu, Zeyu Jiang, Haozhi Cao, Shenghai Yuan, and Lihua Xie are with the Centre for Advanced Robotics Technology Innovation (CARTIN), School of Electrical and Electronic Engineering, Nanyang Technological University, 50 Nanyang Avenue, Singapore 639798, {\tt\small \{kuan.xu, \  zjiang015,haozhi002,shyuan,elhxie\}@ntu.edu.sg}.}
\thanks{$^{2}$Chen~Wang is with Spatial AI \& Robotics Lab, Department of Computer Science and Engineering, University at Buffalo, Buffalo, NY 14260 {\tt\small chenw@sairlab.org}.}
}

\begin{document}

\maketitle
\thispagestyle{empty}

\pagestyle{empty}

\begin{abstract}
    Scene Coordinate Regression (SCR) is a visual localization technique that utilizes deep neural networks (DNN) to directly regress 2D-3D correspondences for camera pose estimation. However, current SCR methods often face challenges in handling repetitive textures and meaningless areas due to their reliance on implicit triangulation.
    In this paper, we propose an efficient and accurate SCR system.
    Compared to existing SCR methods, we propose a unified architecture for both scene encoding and salient keypoint detection, allowing our system to prioritize the encoding of informative regions. This design significantly improves computational efficiency.
    Additionally, we introduce a mechanism that utilizes sequential information during both mapping and relocalization. The proposed method enhances the implicit triangulation, especially in environments with repetitive textures.
    Comprehensive experiments conducted across indoor and outdoor datasets demonstrate that the proposed system outperforms state-of-the-art (SOTA) SCR methods. 
    Our single-frame relocalization mode improves the recall rate of our baseline by 6.4\% and increases the running speed from 56$\hertz$ to 90$\hertz$.
    Furthermore, our sequence-based mode increases the recall rate by 11\% while maintaining the original efficiency.
\end{abstract}



\section{Introduction}

    Visual relocalization is crucial for many applications such as mobile robots and augmented reality, offering a practical balance between cost and accuracy  \cite{brachmann2021visual}. The core task involves estimating the camera pose using a pre-built map. The map can be generally classified into two categories: explicit maps and implicit maps. Explicit maps consist of a 3D point cloud combined with visual descriptors of 2D features \cite{schneider2018maplab, sarlin2019coarse}, while implicit maps encode the scene using a neural network \cite{brachmann2021visual, kendall2017geometric}.
    Feature matching (FM)-based methods construct explicit maps using techniques like structure-from-motion (SFM) \cite{sarlin2019coarse, schoenberger2016sfm} or simultaneous localization and mapping (SLAM) \cite{schneider2018maplab, xu2025airslam}. During relocalization, these methods rely on image retrieval and feature matching to establish 2D-3D correspondences between pixels in the query image and 3D points in the map. However, this approach often suffers from large map sizes due to the storage of visual features and 3D models. Additionally, the process can be computationally inefficient, as the feature matching needs to be performed between the query image and multiple candidates.

    In contrast, scene coordinate regression (SCR) methods encode map information into a neural network \cite{brachmann2021visual, dong2022visual}. They directly regress the 3D coordinates in the world coordinate system correspondences to the image pixels, eliminating the need for explicit 2D-3D corresponding searching. Compared to FM-based methods, SCR methods enable faster relocalization with a smaller map. Recently, ACE \cite{brachmann2023accelerated} proposes a generalized convolutional neural network (CNN) to extract feature maps for scene encoding. Building on this pre-trained backbone, ACE utilizes a compact 4MB-sized multilayer perceptron (MLP) head as the map, enabling new scene encoding in five minutes only without requiring a 3D model or depth information. This approach achieves state-of-the-art accuracy and efficiency.

    \begin{figure}[t]
        \vspace{0.5em}
        \centering
        \setlength{\abovecaptionskip}{0.3cm}
        \includegraphics[width=0.99\linewidth]{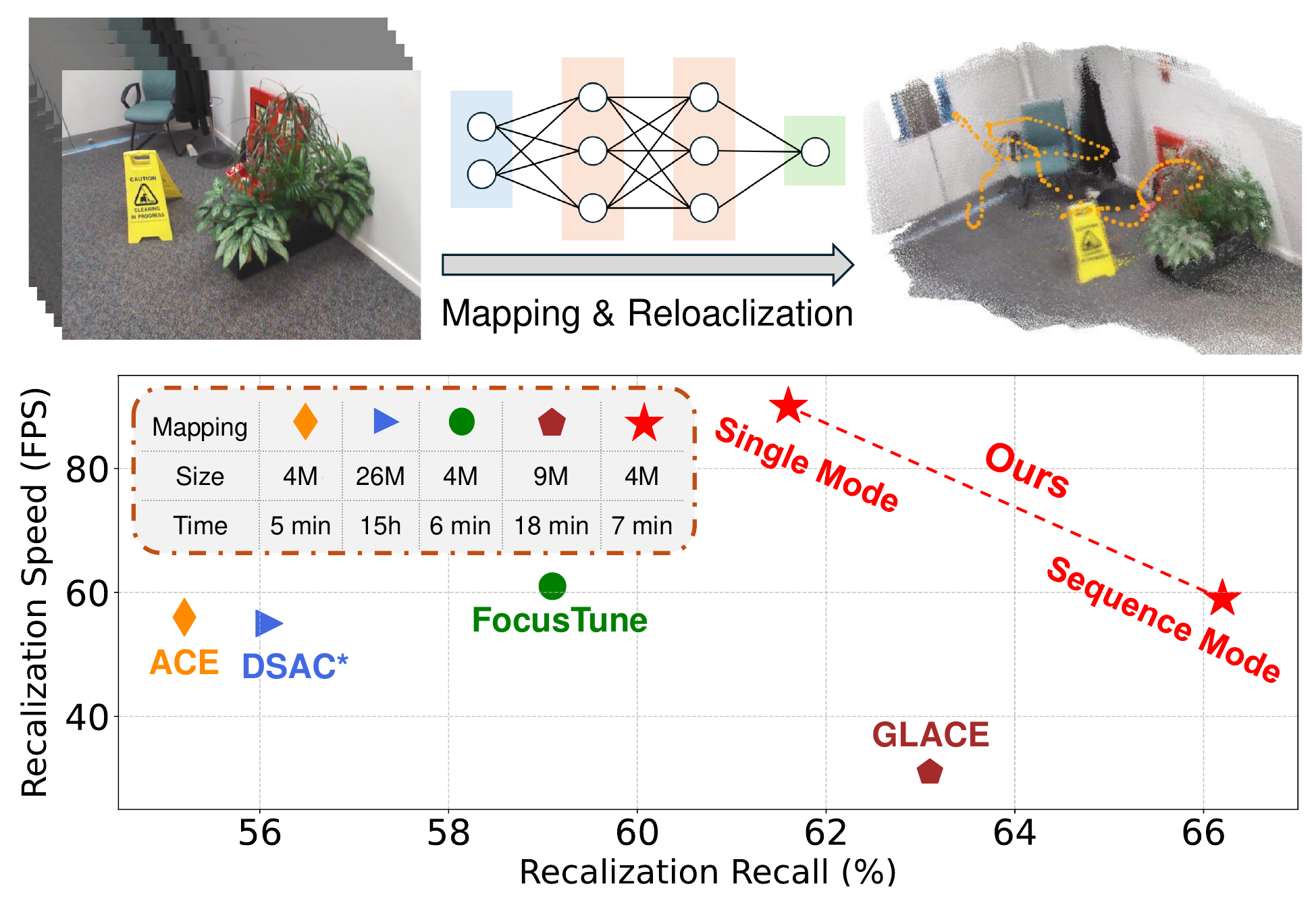}
        \caption{In this paper, we propose an efficient SCR system that leverages deep neural networks for both mapping and relocalization. Our system delivers accurate and efficient relocalization, while maintaining fast mapping and a compact map size, achieving state-of-the-art performance.}
        \label{fig:fig1}
    \end{figure}

    Despite their advantages, SCR methods still face several notable limitations.
    First, the presence of many texture-less and non-informative regions in the image significantly increases the map encoding errors \cite{nguyen2024focustune}.
    Second, existing RGB-only SCR methods rely heavily on implicit triangulation, \ie the neural network must consistently identify and map the same regions across different images to accurate 3D points. This reliance causes a sharp decline in performance in environments with repetitive textures \cite{wang2024glace}.
    Third, during relocalization, all pixel-to-3D matches are considered for pose estimation since the reliability of specific regions is unknown, which significantly reduces efficiency.
    Efforts have been made to address these challenges. FocusTune \cite{nguyen2024focustune} mitigates encoding errors by focusing on informative regions but requires a 3D model of the scene, which limits its practicality. For the second issue, GLACE \cite{wang2024glace} introduces global image descriptors to better distinguish similar regions across images, yet this approach comes at the cost of reduced mapping and localization efficiency.

    In this paper, we present a novel SCR system designed to tackle these challenges. To address the first and third issues, we introduce a unified and scene-agnostic CNN backbone that simultaneously encodes the scene and identifies salient regions, focusing mapping and relocalization efforts exclusively on these informative areas. For the second challenge, we explicitly associate the salient regions across different images, leveraging sequential information to enhance implicit triangulation and relocalization.
    In summary, the main contributions of this work are as follows:
    \begin{itemize}
    \item We introduce a unified, scene-agnostic CNN backbone for scene encoding and salient region detection, which mitigates accuracy degradation caused by non-informative areas.
    \item We propose a novel method that leverages sequential information to strengthen the implicit triangulation in SCR systems, improving mapping accuracy, particularly in environments with repetitive textures.
    \item We design two relocalization modes: a single-frame mode that delivers one-shot relocalization with high efficiency and a sequence mode that utilizes temporal information to enhance relocalization performance.
    \item Extensive experimental results show that our methods achieve substantial improvement in both efficiency and effectiveness. We provide the source code at \textcolor{blue}{\url{https://github.com/sair-lab/SeqACE}}.
    \end{itemize}

\section{Related Works}

\subsection{Feature Matching Based Relocalization}\label{sec:related-fm}

FM-based visual relocalization methods remain the state-of-the-art for large-scale and challenging environments \cite{wang2024glace}. These approaches rely on 3D models generated through SfM or SLAM \cite{schneider2018maplab, sarlin2019coarse, zhang2023map, liu2024pe}. During relocalization, candidate map images similar to the query image are retrieved using techniques such as bag-of-words (BoW) or image retrieval methods \cite{sarlin2019coarse, xu2025airslam}. Feature matching is then performed between the query image and each retrieved map image. Since the keypoints in the map images have already been triangulated, 2D-3D correspondences between the pixels in the query image and the 3D points in the point cloud model can be established. Finally, the 6-DOF pose of the query image is estimated using geometric techniques, such as Perspective-n-Point (PnP) and bundle adjustment \cite{zhou2022geometry, sarlin2021back}.


\subsection{Absolute Pose Regression}\label{sec:related-apr}

Unlike FM-based methods, absolute pose regression (APR) approaches use a DNN to regress camera poses from RGB images directly. They have also garnered notable attention recently due to their good efficiency and simple architecture. PoseNet \cite{kendall2015posenet} is the first APR system. The subsequent works explore different architectural designs \cite{melekhov2017image} or leverage sequential information \cite{clark2017vidloc} to enhance APR systems. Recent APR works utilize novel synthesis techniques to generate high-quality training data \cite{purkait2018synthetic} or integrate them for fine-tuning \cite{chen2021direct}. However, APR approaches still fall short in accuracy compared to SCR and FM-based methods.

\subsection{Scene Coordinate Regression}\label{sec:related-scr}
Different from FM-based methods and APR approaches, SCR methods regress the corresponding 3D coordinates in the scene for 2D pixels in the query image. Most existing SCR methods require depth information or a 3D model of the scene \cite{cai2019camera, tang2021learning}.  
Dong \etal \cite{dong2022visual} use a pre-trained feature extractor and scene classifier for efficient few-shot mapping, but extra modules limit localization speed to $5\hertz$ even on a high-end GPU. Zhou \etal \cite{zhou2020kfnet} introduce KFNet, integrating a Kalman filter \cite{kalman1960new} into the SCR pipeline to leverage sequential information. Bui \etal \cite{bui2024d2s} propose D2S, which maps sparse feature descriptors to 3D scene points using a pre-trained network.

DSAC++ \cite{brachmann2018learning} is the first SCR method independent of depth or 3D models but requires long training. DSAC* \cite{brachmann2021visual} streamlines DSAC++ by unifying initialization steps, cutting training time from 6 days to 15 hours. ACE \cite{brachmann2023accelerated} further improves efficiency by using a pre-trained CNN, reducing mapping time to 4 minutes and map size to 4MB.
FocusTune \cite{nguyen2024focustune}, GLACE \cite{wang2024glace}, and EGFS \cite{liu2024reprojection} are all variants of ACE, each introducing unique enhancements. FocusTune refines ACE by projecting a 3D model onto images to filter irrelevant regions, but it depends on the 3D model. EGFS refines region selection through an iterative process using a segmentation model and reprojection errors. Meanwhile, GLACE enhances ACE by incorporating global image descriptors for better region distinction.
Our system also builds on ACE. However, unlike these methods, we utilize keypoint detection to filter out uninformative regions and leverage sequential information to reduce visual aliasing.

\subsection{Keypoint Detection}\label{sec:related-kp}
Keypoint detection is essential for visual localization. Traditional methods like FAST \cite{viswanathan2009features} and ORB \cite{rublee2011orb} rely on handcrafted detectors and descriptors, offering efficiency but often lacking robustness. In contrast, learning-based approaches, such as SuperPoint \cite{detone2018superpoint} and R2D2 \cite{revaud2019r2d2}, leverage deep networks to enhance repeatability and precision. Recently, detector-free methods like D2-Net \cite{dusmanu2019d2} jointly learn detection and description, improving performance in challenging scenarios.  To optimize efficiency, keypoint detection can leverage a shared backbone network in multi-task learning, enabling integration with tasks like local-global feature extraction \cite{sarlin2019coarse} or joint point-line detection \cite{xu2025airslam}. Inspired by \cite{sarlin2019coarse, xu2025airslam}, we use a shared backbone for both SCR and keypoint detection, improving system efficiency.

\section{Preliminaries}

\subsection{Problem Statement}\label{sec:preliminaries-ps}
The objective is to estimate the pose $\mathbf{T}_{wc}$ of a given image $I$, where $\mathbf{T}_{wc}$ denotes the transformation from the camera coordinate system to the world coordinate system. Given a set of 2D-3D correspondences $\mathcal{C} = {(\mathbf{p}_i, \mathbf{P}_i^w)}$, where $\mathbf{p}_i$ represents a pixel in the image $I$ and $\mathbf{P}_i^w$ is the corresponding 3D point in the world coordinate system, the pose can be efficiently computed using PnP algorithms or iterative optimization techniques. To obtain these correspondences $\mathcal{C}$, SCR methods aim to learn a mapping from 2D pixels to their corresponding 3D points:
\begin{equation}\label{eq:scr_encoding}
   \mathbf{P}_i^w = f \left( \mathcal{P} \left( \mathbf{p}_i, I \right); \mathbf{w} \right),
\end{equation}
where $\mathcal{P} \left( \mathbf{p}_i, I \right) \in \mathbb{R}^{{C_p} \times {H_p} \times {W_p}} $ is a small patch centered at $\mathbf{p}_i$ in the image $I$, and $f$ is a neural network with learnable parameters $\mathbf{w}$. Existing SCR pipelines usually take grayscale inputs with $C_p = 1$. As there is no explicit patch extraction, the patch size is determined by the receptive field of the neural network. Considering the trade-off between invariance and discriminative power, the patch size is typically set to $H_p = W_p = 81\text{px}$ in many existing architectures \cite{brachmann2023accelerated, wang2024glace}.

\subsection{Accelerated Coordinate Encoding}\label{sec:preliminaries-ace}

Training $f$ in \equref{eq:scr_encoding} using only posed images takes a long time \cite{brachmann2021visual}. To accelerate this process, Accelerated Coordinate Encoding (ACE) \cite{brachmann2023accelerated} proposes to split $f$ into a scene-agnostic backbone $f_B$ and a scene-specific MLP head $f_H$:
\begin{equation}\label{eq:ace_split}
        \mathbf{P}_i^w = f_H \left( \mathbf{f}_i ; \mathbf{w}_H \right), \; \text{with}  \ \mathbf{f}_i = f_B \left( \mathcal{P} \left( \mathbf{p}_i, I \right); \mathbf{w}_B \right), 
\end{equation}
where $f_B$ predicts a high-dimensional feature vector $\mathbf{f}_i$ for each patch, and $f_H$ maps $\mathbf{f}_i$ to the corresponding 3D point:
\begin{equation}\label{eq:ace_mapping}
        f_B: \mathbb{R}^{{C_p} \times {H_p} \times {W_p}} \xrightarrow{} \mathbb{R}^{C_{\mathbf{f}}}, \; \text{and}  \  f_H: \mathbb{R}^{C_{\mathbf{f}}} \xrightarrow{} \mathbb{R}^{3}, 
\end{equation}
where $C_{\mathbf{f}}$ is the dimension of $\mathbf{f}_i$.
The backbone $f_B$ is pre-trained on a large dataset. When mapping a new scene, only the MLP head $f_H$ needs to be trained while $f_B$ remains fixed. Thus, the MLP head serves as the map. This approach significantly reduces both the map size and the mapping time.

\begin{figure*}[t]
    \vspace{0.5em}
    \centering
    \setlength{\abovecaptionskip}{0.2cm}
    \includegraphics[width=0.98\linewidth]{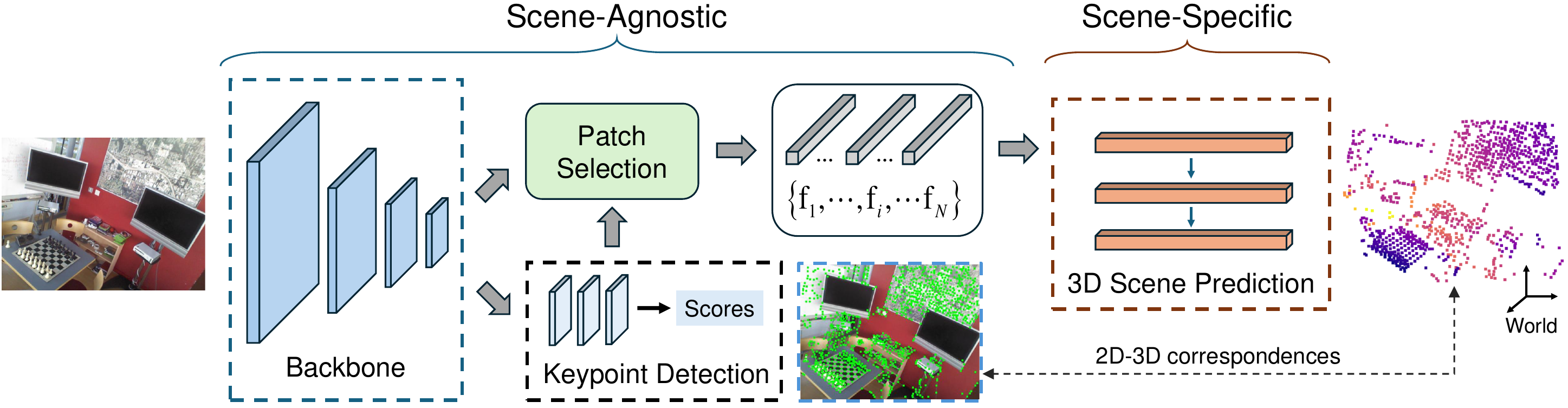}
    \caption{Our system comprises a scene-agnostic module and a scene-specific module. The scene-agnostic module identifies informative image patches and generates high-dimensional features for each patch. The scene-specific module, implemented using 1D convolutional layers, predicts the corresponding 3D scene point for each patch. These 2D-3D correspondences are then utilized for relocalization.
    }
    \label{fig:overview}
    \vspace{-0.5em}
\end{figure*}

\section{Methods}

\subsection{Overview}\label{sec:overview}
The overall structure of the proposed system is depicted in \fref{fig:overview}. Our system consists of a scene-agnostic module and a scene-specific module. The scene-agnostic module generates feature vectors and saliency scores for image patches, while the scene-specific module, designed as an MLP head with several 1D convolutional layers, maps these feature vectors to 3D scene points.  
The detailed design of the scene-agnostic module is presented in \sref{sec:scm}. \sref{sec:mapping} describes how sequential information is utilized to enhance the training of the scene-specific module, and \sref{sec:reloc} outlines our relocalization methods.

\subsection{Scene-Agnostic Module}\label{sec:scm}
In this section, we aim to design an efficient and scene-agnostic module capable of (1) detecting informative image patches and (2) mapping these patches into high-dimensional features for scene encoding. As many learning-based patch and keypoint detection methods have been proposed \cite{sarlin2019coarse, detone2018superpoint}, a straightforward solution is to utilize them directly for salient region detection along with the ACE backbone. However, employing two separate deep neural networks would significantly reduce online efficiency. 
Inspired by \cite{sarlin2019coarse, xu2025airslam}, we accomplish these two tasks with a shared backbone.
To achieve this, we extend the ACE backbone by adding a lightweight keypoint detection head (KDH). The KDH consists of just three convolutional layers, introducing trivial computational overhead. 
The KDH takes the feature map from the backbone as input and outputs a tensor $\mathcal{T}_K \in \mathbb{R}^{65 \times \frac{H}{8} \times \frac{W}{8}}$, where $H$ and $W$ denote the height and width of the input image, respectively.
The 65 channels correspond to an $8 \times 8$ grid region and a dustbin channel representing the absence of keypoints. This tensor is processed through a softmax layer and resized into a $H \times W$ heatmap, which scores the saliency of the corresponding image pixels.
Since the KDH, similar to the backbone \( f_B \) in \equref{eq:ace_split}, is scene-agnostic, we integrate them into a unified module, denoted as \( f_S \), with parameters represented by \( \mathbf{w}_S \).
Then the second equation in \equref{eq:ace_split} is updated to:
\begin{equation}\label{eq:sqr_backbone}
    \mathbf{f}_i, \mathcal{S}_i = f_S \left( \mathcal{P} \left( \mathbf{p}_i, I \right); \mathbf{w}_S \right), 
\end{equation}
where $\mathcal{S}_i$ is the salient score of $\mathbf{p}_i$. 

We train \( f_S \) using a two-step pipeline on the ScanNet dataset \cite{dai2017scannet}. In the first step, we follow the approach outlined in \cite{brachmann2023accelerated} to train the backbone \( f_B \). In the second step, the parameters of \( f_B \) are fixed, and the distillation techniques are utilized to train the KDH.
We use the prediction of SuperPoint \cite{detone2018superpoint} as supervision signals as it is simple, efficient, and performs excellently. The loss function is defined as:
\begin{equation}\label{eq:kdh_loss}
    \mathcal{L}_K = \mathcal{L}_{CE} \left( \mathcal{T}_K, \mathcal{T}'_K \right), 
\end{equation}
where $\mathcal{L}_{CE}$ is the cross-entropy loss and $\mathcal{T}'_K$ is the prediction of the teacher network.
We also experimented with a multi-task learning approach \cite{sarlin2019coarse} to train $f_S$, but found that the two-step pipeline produced better results. 
This may be attributed to the substantial difference in difficulty between these two tasks. Specifically, training a keypoint detector converges much faster than training a universal scene encoder. This disparity in task difficulty may lead to suboptimal training within a multitask network \cite{chen2018gradnorm}.

\begin{figure}[t]
    \vspace{0.5em}
    \centering
    \setlength{\abovecaptionskip}{0.3cm}
    \includegraphics[width=0.99\linewidth]{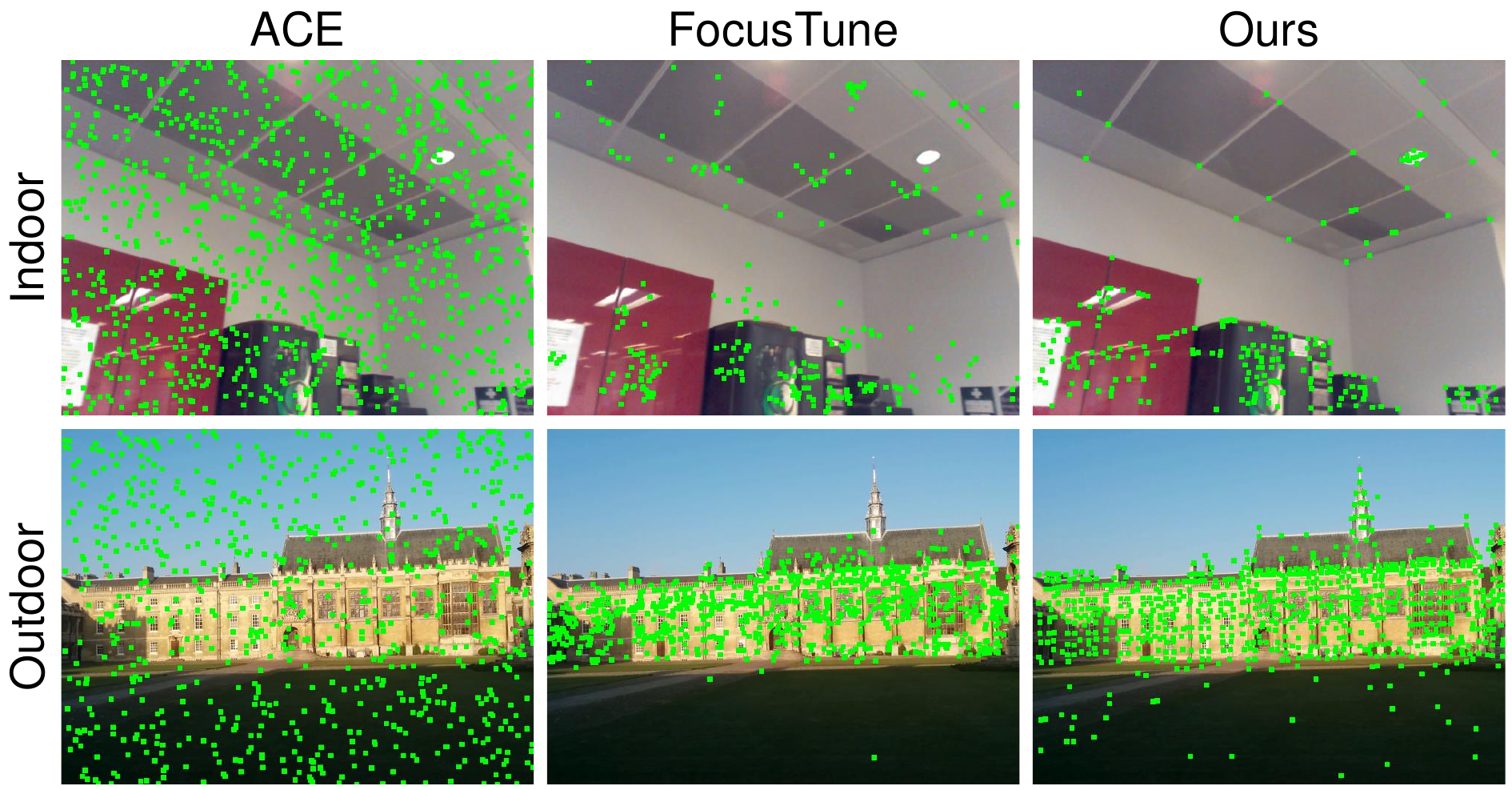}
    \caption{Sampling strategies comparison. ACE \cite{brachmann2023accelerated} randomly samples image patches, while FocusTune \cite{nguyen2024focustune} performs uniform sampling around the projected points of the 3D model. In contrast, we employ keypoint detection to identify repeatable image patches.}
    \label{fig:sampling}
\end{figure}

\subsection{Sequence-Based Mapping}\label{sec:mapping}

\paragraph{Saliency-Based Buffer Sampling} 
Buffer sampling is the key to fast mapping. Since the backbone, \ie $f_B$ or $f_S$, is scene-agnostic, it only needs to be executed once when mapping a new scene. The resulting outputs, $\{\mathbf{f}_i\}$, can be cached and subsequently used to train $f_H$. To enhance efficiency, not all image patches are utilized during training.
In ACE \cite{brachmann2023accelerated}, 1024 image patches are randomly sampled per image, and their corresponding $\mathbf{f}_i$, $\mathbf{p}_i$, camera intrinsics, and camera poses are stored in a training buffer. However, this strategy may sample many meaningless regions. To address this, FocusTune \cite{nguyen2024focustune} projects the 3D point cloud of the scene onto the training images and performs uniform sampling around the projected points. While this approach better targets informative areas, FocusTune cannot guarantee consistent sampling across multiple images.

In contrast, we utilize the keypoint detection results from \sref{sec:scm} to select the top 1,000 patches with the highest confidence scores in each image, aligning with the number used by ACE. This approach guarantees consistent patch selection across image sequences, and also ensures the same patches are utilized during both mapping and relocalization.
We show a comparison of patch selection in \fref{fig:sampling}.

\paragraph{Self-Loss}
We follow the previous works \cite{brachmann2023accelerated, wang2024glace} to compute a reprojection loss to supervise the training of $f_H$:
\begin{equation}\label{eq:self_loss}
    \mathcal{L}_{S, i} = \left\{ \begin{array}{lcl}
        \tau (t) \tanh \left( \frac{e_\pi (\mathbf{p}_i, \mathbf{P}_i^w, \mathbf{T}_{wc})}{\tau(t)} \right) & \mbox{if}  \; \mathbf{P}_i^w \in \mathcal{V} \\
        \lVert \mathbf{P}_i^w - \overline{\mathbf{P}}_i^w \lVert & \mbox{otherwise} ,
        \end{array}\right.
\end{equation}
where $\mathbf{P}_i^w$ is the 3D scene point predicted by $f_H$ using $\mathbf{f}_i$ and $e_\pi (\mathbf{p}_i, \mathbf{P}_i^w, \mathbf{T}_{wc})$ is the reprojection error. $\mathcal{V}$ is a set of valid predictions that are between 0.1m and 1000m in front of the image plane and have a reprojection error of less than 1000px. $\scriptstyle{\overline{\mathbf{P}}_i^w}$ is the pseudo ground truth 3D point computed by the inverse projection of $\mathbf{p}_i$ with the camera pose $\mathbf{T}_{wc}$ and a fixed target depth at 10m. $\tau (t)$ is a dynamically adjusted scale between $\tau_{\min} = 1$ and $\tau_{\max} = 50$:
\begin{equation}\label{eq:loss_tau}
    \tau (t) = \tau_{\max} \sqrt{1 - t^2} + \tau_{\min},
\end{equation}
where $t \in (0, 1)$ represents the relative training progress.


For each image patch, $\mathcal{L}_{S, i}$ supervises the prediction of $\mathbf{P}_i^w$ for each $\mathbf{f}_i$ using only a single image, which is why we refer to it as self-loss.
However, in classical multi-view geometry, triangulating a point successfully requires at least two observations \cite{wang2024glace, hartley2003multiple}.
Therefore, SCR systems heavily rely on implicit triangulation, implying that the neural network must consistently recognize the same region across multiple images to ensure that scene points are constrained by observations from different viewpoints. This is particularly challenging under complex environments.

\paragraph{Cross-Loss}
To tackle the above challenge, we propose explicitly integrating multi-view constraints into the mapping process, rather than relying solely on the neural network to learn them implicitly. The core idea is to associate the selected patches across different images using feature-matching techniques \cite{lindenberger2023lightglue}. This approach enables us to project a predicted 3D point onto multiple frames and leverage the resulting reprojection errors for more effective supervision.

Inspired by visual SLAM techniques, we select keyframes to ensure that matching patch pairs have varying baselines. A frame is chosen as a keyframe if the matched features with the previous keyframe are less than half of the detected features. For each matching pair $(\mathbf{p}_{k, j}, \mathbf{p}_{i})$, where $\mathbf{p}_{k, j}$ is the matched point on the keyframe, we define a loss as:
\begin{equation}\label{eq:cross_loss}
    \mathcal{L}_{C, i} = \left\{ \begin{array}{lcl}
        \tau (t) \tanh \left( \frac{e_\pi (\mathbf{p}_{k, j}, \mathbf{P}_i^w, \mathbf{T}_{wk})}{\tau(t)} \right) & \mbox{if}  \; \mathbf{P}_i^w \in \mathcal{V} \\
        \lVert \mathbf{P}_i^w - \overline{\mathbf{P}}_{k, j}^w \lVert & \mbox{otherwise} ,
        \end{array}\right.
\end{equation}
where $\mathbf{T}_{wk}$ is the pose of the keyframe and $\overline{\mathbf{P}}_{k, j}^w$ is the pseudo ground truth 3D point computed by the inverse projection of $\mathbf{p}_{k, j}$. Since $\mathcal{L}_{C, i}$ uses different frames to supervise the prediction of scene points, we call it cross-loss. 

\paragraph{Training Details}
For each image patch, we store the corresponding feature vector $\mathbf{f}_i$, the pixel $\mathbf{p}_i$, the camera intrinsic $\mathbf{K}$, the camera pose $\mathbf{T}_{wc}$, the pixel matched on the keyframe $\mathbf{p}_{k, j}$, and the keyframe pose $\mathbf{T}_{wk}$, in the training buffer. We follow ACE \cite{brachmann2023accelerated} to shuffle these patches and select $\mathcal{N}_b = 5120$ patches in each iteration to compute the loss:
\begin{equation}\label{eq:sum_loss}
    \mathcal{L} = \frac{1}{\mathcal{N}_b} \sum \left( \mathcal{L}_{S, i} + \lambda \mathcal{L}_{C, i} \right),
\end{equation}
where $\lambda = 0.5$ if $(\mathbf{p}_{k, j}, \mathbf{p}_{i})$ is an inlier match, otherwise $\lambda = 0$. We train $f_H$ over 16 complete iterations using the AdamW optimizer \cite{loshchilov2018decoupled} with a learning rate between $5e\text{-}4$ and $5e\text{-}3$ and a 1-cycle schedule \cite{smith2019super}. 

\begin{table*}[t]
    \caption{Comparison on the 7-Scenes dataset. The relocalization recall is the proportion of test frames below a (1cm, 1$^\circ$) pose error.}
    \vspace{-1pt}
    \label{tab:7scenes_recall}
    \centering
    \scriptsize 
    \begin{threeparttable}
    \resizebox{0.98\linewidth}{!}{
    \begin{tabular}{C{0.03\linewidth}|C{0.13\linewidth}|C{0.06\linewidth}|C{0.05\linewidth}|C{0.03\linewidth}|C{0.03\linewidth}C{0.025\linewidth}C{0.03\linewidth}C{0.03\linewidth}C{0.045\linewidth}C{0.04\linewidth}C{0.03\linewidth}>{\columncolor{gray!20}}C{0.03\linewidth}}
        \toprule
        \multirow{2}{*}{} & \multirow{2}{*}{} & Mapping & Map & \multirow{2}{*}{FPS\tnote{*}} &  \multicolumn{8}{c}{Relocalization Recall (\%)} \\
         & & Time & Size  & & Chess  & Fire & Heads & Office & Pumpkin & Kitchen & Stairs & Avg.\\
         \midrule  
         \multirow{3}{*}{FM}
        &  GoMatch \cite{zhou2022geometry} & \cellcolor{red!20}1.5h & \cellcolor{orange!30}$\sim$300M & \cellcolor{red!20}1.23 & 1.6 & 1.4 & 0 & 0.6 & 0.3 & 0.5 & 0.7 & 0.7 \\
        &  PixLoc \cite{sarlin2021back} & \cellcolor{red!20}1.5h & \cellcolor{red!20}$\sim$1.7G & \cellcolor{red!20}1.23 & 89.1 & 66.2 & 77.6 & 55.3 & 56.7 & 66.5 & 9.8 & 60.2 \\
        & HLoc (SP+SG) \cite{sarlin2019coarse} & \cellcolor{red!20}1.5h & \cellcolor{red!20}$\sim$3.5G & \cellcolor{red!20}0.46 & 96.1 & 71.2 & 82.5 & 68.6 & 75.2 & 79 & 16.3 & 69.8 \\
         \midrule  
         \multirow{6}{*}{SCR}
        & DSAC* \cite{brachmann2021visual} & \cellcolor{red!20}9h & \cellcolor{orange!30}26.3M & \cellcolor{orange!30}55 & 82.3 & 59.3 & 85.4 & 39.4 & 56.4 & 67.6 & 2.6 & 56.1 \\
        & ACE \cite{brachmann2023accelerated} & \cellcolor{green!20}5 min & \cellcolor{green!20}4.1M & \cellcolor{orange!30}56 & 87.2 & 60.1 & 87.9 & 46.2 & 34.3 & 66.7 & 3.8 & 55.2 \\
        & FocusTune \cite{nguyen2024focustune} & \cellcolor{green!20}6 min & \cellcolor{green!20}4.1M & \cellcolor{orange!30}61 & 90.4 & 64.3 & 90.1 & 49.6 & 47.1 & 68.3 & 3.9 & 59.1  \\
        & GLACE \cite{wang2024glace} & \cellcolor{orange!30}18 min & \cellcolor{green!20}9M & \cellcolor{orange!30}31 & 87.4 & 67.1 & 86.3 & \textbf{60.5} & \textbf{58.7} & \textbf{75.8} & 5.8 & 63.1 \\
        & Ours-Single & \cellcolor{green!20}7 min & \cellcolor{green!20}4.1M & \cellcolor{green!20}90 & 94.2 & 63.5 & 87.9 & 56.2 & 51.9 & 67.7 & 11.0 & 61.6 \\
        & Ours-Sequence & \cellcolor{green!20}7 min & \cellcolor{green!20}4.1M & \cellcolor{orange!30}59 & \textbf{96.5} & \textbf{68.8} & \textbf{91.7} & 59.8 & 57.8 & 71.7 & \textbf{17.4} & \textbf{66.2} \\
         \bottomrule

    \end{tabular}
    }
        \begin{tablenotes}
            \footnotesize
            \item[*] Running time of relocalization measured in frame per second (FPS).
        \end{tablenotes}
    \end{threeparttable}
    \vspace{-2.0em}
\end{table*}

\subsection{Relocalization}\label{sec:reloc}

\subsubsection{Single-Mode}\label{sec:sigle-reloc}
Our system supports two relocalization modes. The single-mode estimates the camera pose from a single image, offering efficiency and simplicity. As shown in \fref{fig:overview}, it detects 2D keypoints, predicts corresponding 3D scene points, and applies PnP with RANSAC to optimize the pose by maximizing inliers.
Note that this mode also uses the map built with our sequence-based mapping method.

\subsubsection{Sequence-Mode}\label{sec:seq-reloc}
Like most existing SCR methods, our single-mode does not leverage the constraints between adjacent frames in an image sequence during relocalization. However, in real-world robotic applications, the previous frame provides valuable prior information for pose estimation. Therefore, we also design a sequence-mode that utilizes sequential information to enhance the relocalization.
Details of this mode are in the following section.

\begin{figure}[t]
    \vspace{0.5em}
    \centering
    \setlength{\abovecaptionskip}{0.1cm}
    \includegraphics[width=0.95\linewidth]{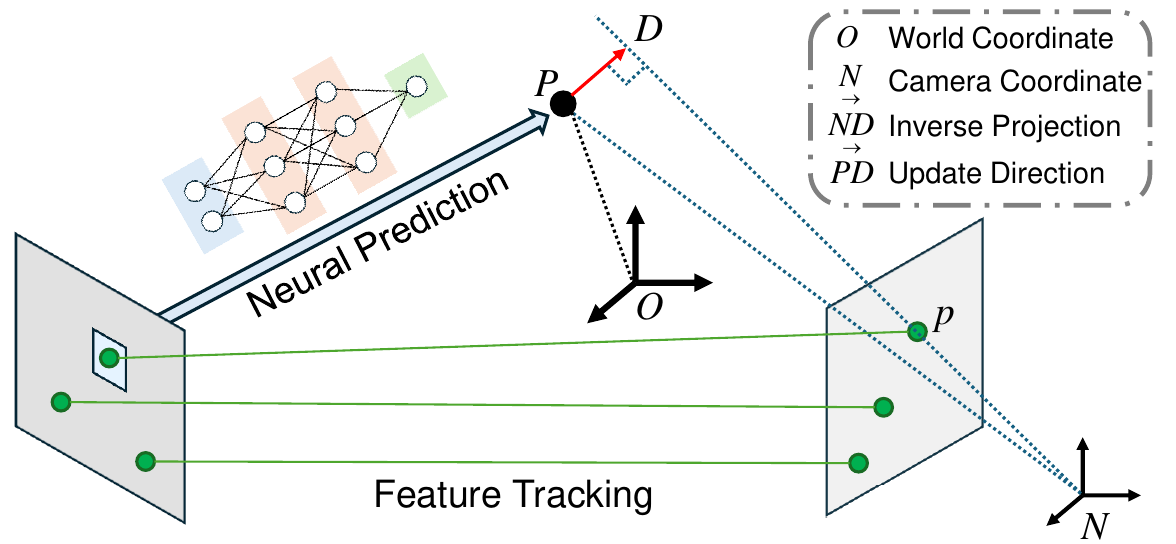}
    \caption{In our sequence-based relocalization, the predicted 3D scene points are used to constrain the pose estimation of subsequent frames and updated with new observations.}
    \label{fig:scene_update}
    \vspace{-0.5em}
\end{figure}

\paragraph{Pose Prediction}
In traditional visual odometry methods \cite{xu2023airvo}, local constraints are established by first triangulating 3D points from historical frames and then using these points to constrain the current pose. Typically, triangulation involves solving an $\mathbf{A} \mathbf{x} = \mathbf{b}$ problem through matrix decomposition, which can be computationally intensive. To simplify this process, we propose a more efficient approach. 
Our key insight is that if the reprojection errors of a predicted scene point across multiple historical frames remain small, the prediction can be considered accurate and used to constrain the current pose. Based on this idea, we maintain a set of reliably predicted scene points for pose estimation.

We denote two adjacent frames as $\mathcal{I}_m$ and $\mathcal{I}_n$, where $\mathcal{I}_n$ is the current frame and $\mathcal{I}_m$ is the previous frame. Let $\mathcal{Y}_m$ and $\mathcal{X}_m$ represent the maintained set of 3D scene points and their corresponding 2D points on $\mathcal{I}_m$, respectively. Using optical flow \cite{lucas1981iterative}, we track $\mathcal{X}_m$ to obtain $\mathcal{\bar{X}}_n$ on $\mathcal{I}_n$, along with their corresponding scene points $\mathcal{\bar{Y}}_n$ (denoted with a bar to indicate pre-update values). The pose of $\mathcal{I}_n$ is then computed as: 
\begin{equation}\label{eq:pose_prediction}
    \mathbf{\bar{T}}_{wn} = \mathfrak{g} \left(\mathcal{\bar{X}}_n,  \mathcal{\bar{Y}}_n \right),
\end{equation}
where $\mathfrak{g}$ represents the PnP solver with RANSAC.

\paragraph{Pose Update}
The pose estimation in \equref{eq:pose_prediction} only uses 2D-3D correspondences of historical frames and feature tracking results, so $\mathbf{\bar{T}}_{wn}$ can be refined with new predictions from the network for $\mathcal{I}_n$. 
Let $\mathcal{\hat{X}}_n$ and $\mathcal{\hat{Z}}_n$ denote the newly detected keypoints and the corresponding 3D scene points predicted by the network. Since the set of 2D-3D correspondences $\{\mathcal{\hat{X}}_n, \mathcal{\hat{Z}}_n\}$ typically contains many outliers, we first use PnP with RANSAC to compute a pose $\mathbf{\hat{T}}_{wn} = \mathfrak{g} \left(\mathcal{\hat{X}}_n,  \mathcal{\hat{Z}}_n \right)$, then fuse $\mathbf{\hat{T}}_{wn}$ with $\mathbf{\bar{T}}_{wn}$ at the pose level, rather than directly using a Kalman filter. We perform a simple weighted average on the manifold to obtain the updated pose:
\begin{equation}\label{eq:pose_update}
    \mathbf{T}_{wn} = \exp \left( \left( \bar{w}_n \log(\mathbf{\bar{T}}_{wn})^{\vee} + \hat{w}_n \log (\mathbf{\hat{T}}_{wn})^{\vee} \right) ^{\wedge} \right),
\end{equation}
where $\exp$ and $\log$ represent the exponential and logarithmic maps, respectively. The weights $\bar{w}_n$ and $\hat{w}_n$ are computed as:
\begin{equation}\label{eq:pose_weight}
    \bar{w}_n = \frac{\mathcal{\bar{N}}_n}{\mathcal{\bar{N}}_n + \mathcal{\hat{N}}_n},  \quad   \hat{w}_n = \frac{\mathcal{\hat{N}}_n}{\mathcal{\bar{N}}_n + \mathcal{\hat{N}}_n}
\end{equation}
where $\mathcal{\bar{N}}_n$ and $\mathcal{\hat{N}}_n$ are the number of inliers when estimating $\mathbf{\bar{T}}_{wn}$ and $\mathbf{\hat{T}}_{wn}$ using $\mathfrak{g}$.

\paragraph{Scene Point Update}
The maintained scene point set also requires updating based on new observations from $\mathcal{I}_n$. Besides adding newly detected points and their corresponding predicted 3D points to $ \{\mathcal{\bar{X}}_n, \mathcal{\bar{Y}}_n \}$, we also update tracked scene points. For a scene point $\mathbf{\bar{P}}^w_i \in \mathcal{\bar{Y}}_n$ and the corresponding tracked 2D point $\mathbf{\bar{p}}_i^n \in \mathcal{\bar{X}}_n$, if the pair $(\mathbf{\bar{p}}_i^n, \mathbf{\bar{P}}^w_i)$ is an outlier in \equref{eq:pose_prediction}, it will be removed from $\mathcal{\bar{Y}}_n$ and $\mathcal{\bar{X}}_n$. Otherwise, $\mathbf{\bar{P}}^w_i$ is updated in a direction that reduces the reprojection error on $\mathcal{I}_n$. The update direction is illustrated in \fref{fig:scene_update}, where $\scriptstyle{P}$ and $\scriptstyle{p}$ represent $\mathbf{\bar{P}}^w_i$ and $\mathbf{\bar{p}}_i^n$ for simplicity. The inverse projection line of $\mathbf{\bar{p}}_i^n$ is denoted as $\scriptstyle{\overrightarrow{ND}}$, with $\scriptstyle{D}$ as the foot of the perpendicular. The coordinates of $\scriptstyle{D}$ are: 
\begin{subequations}\label{eq:scene_update_eq}
    \begin{align}
        \mathbf{\bar{D}}^w_i &= \left( \left( \mathbf{\bar{P}}^w_i - \mathbf{t}_{wn}\right)^\top \mathbf{v}_{nd} \right) \mathbf{v}_{nd} - \left( \mathbf{\bar{P}}^w_i - \mathbf{t}_{wn}\right) ,\\
        \mathbf{v}_{nd} &= \mathbf{R}_{wn} \frac{\mathbf{K}_n^{-1} \mathbf{\bar{p}}_i^n}{\| \mathbf{K}_n^{-1} \mathbf{\bar{p}}_i^n \|},
    \end{align}
\end{subequations}
where $\mathbf{R}_{wn}$ and $\mathbf{t}_{wn}$ are the rotation and translation components of $\mathbf{T}_{wn}$, respectively. $\mathbf{v}_{nd}$ is the direction vector of $\scriptstyle{\overrightarrow{ND}}$. Then $\mathbf{\bar{P}}^w_i$ can be updated through:
\begin{equation}\label{eq:point_weight}
    \mathbf{P}^w_i = \mathbf{\bar{P}}^w_i + \beta_n \left( \mathbf{\bar{D}}^w_i - \mathbf{\bar{P}}^w_i \right),
\end{equation}
where $\beta_n$ is a scale factor. As $\mathbf{P}^w_i$ is observed across more frames, the influence of each new observation should diminish, allowing $\mathbf{P}^w_i$ to stabilize gradually. Hence, we set $\beta_n = 1/N_{P}$, where $N_{P}$ denotes the number of observations.

\begin{figure}[t]
    \vspace{0.5em}
    \centering
    \setlength{\abovecaptionskip}{0.3cm}
    \includegraphics[width=0.95\linewidth]{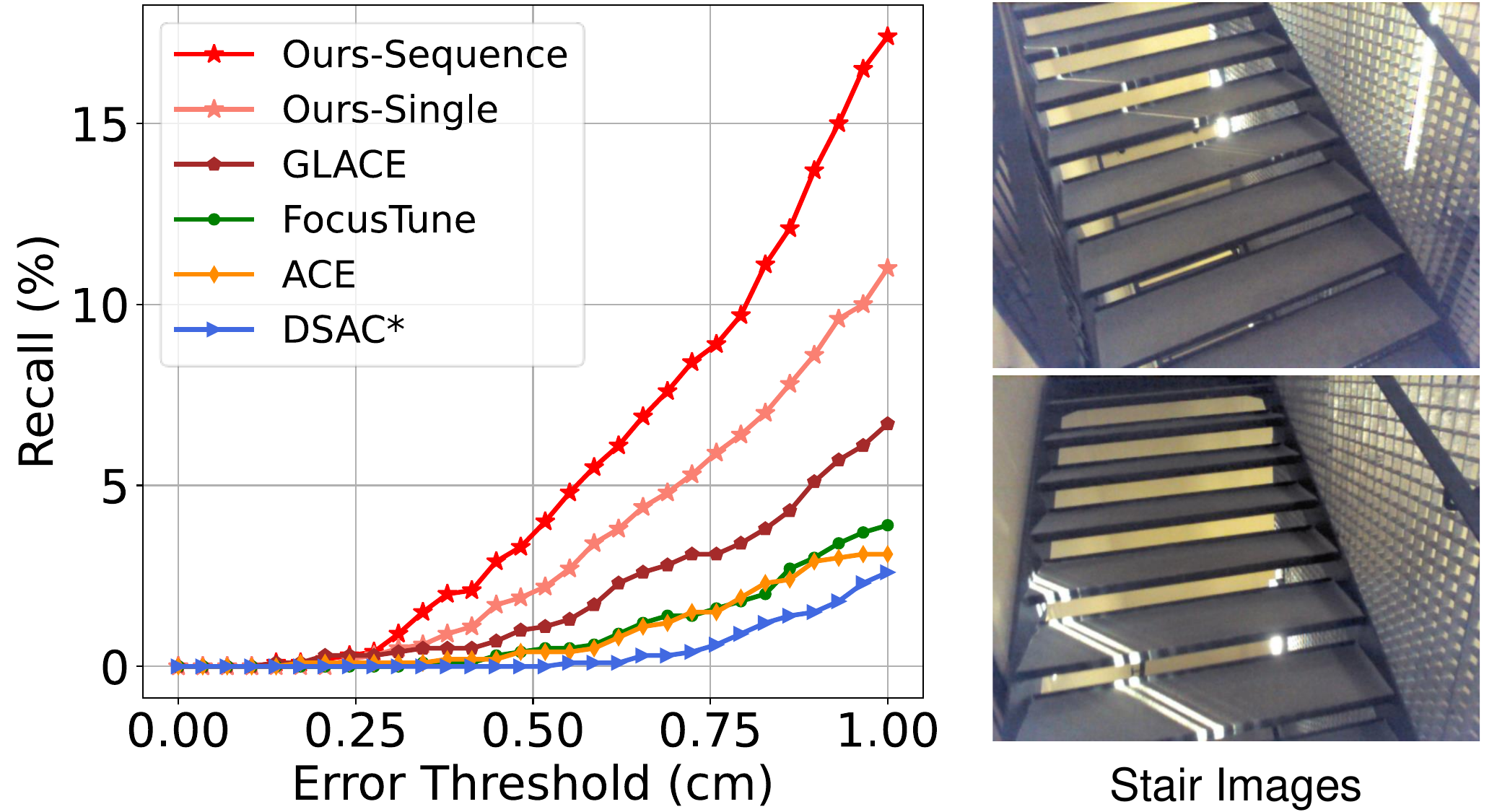}
    \caption{The recall rates in repetitive texture environments as the threshold varies from 0 to 1cm (left) and the scene images (right).
    }
    \label{fig:stairs}
    \vspace{-0.5em}
\end{figure}

\section{Experimental Results} \label{sec:result}
In this section, we present the experimental results. The experiments are performed on three standard datasets for SCR tasks: 7-Scenes \cite{shotton2013scene}, 12-Scenes \cite{valentin2016learning}, and Cambridge Landmarks \cite{kendall2015posenet}. These datasets provide training data for mapping and test data for relocalization. We evaluate the two relocalization modes introduced in \sref{sec:reloc}. In the following sections, ``Ours-Single'' refers to the single-mode relocalization, while ``Ours-Sequence'' or ``Ours'' refers to the sequence-based relocalization. All experiments are conducted on an NVIDIA GeForce RTX 3090 GPU.

\subsection{Comparison with State-of-the-Art}\label{sec:accuracy}

\paragraph{7-Scenes}
The 7-Scenes dataset captures $640 \times 480$ resolution image sequences in seven indoor scenes.
We train $f_H$ using the RGB frames and their ground-truth poses. Our approach is evaluated against several SOTA SCR systems and FM-based methods. Key metrics, including mapping time, map size, relocalization speed, and recall rate, are evaluated. 
The recall rate is defined as the percentage of test frames with pose errors below (1cm, 1$^\circ$). Results are summarized in \tref{tab:7scenes_recall}. 
Our single-mode relocalization achieves remarkable efficiency, running at 90$\hertz$, which is 1.6$\times$ faster than ACE, while also improving the recall by 6.4\%. The sequence mode further delivers SOTA accuracy with high efficiency. By comparison, FocusTune and GLACE, both variants of ACE, have notable drawbacks: FocusTune relies on a 3D model and achieves only a modest accuracy gain (+3.9\%), while GLACE sacrifices mapping and relocalization efficiency. In contrast, our system achieves a substantial recall improvement (+11\%), while preserving ACE's fast mapping time, compact map size, and relocalization speed.

The proposed method significantly enhances system performance in environments with repetitive textures. The ``Stairs'' sequence is captured on a staircase with numerous repetitive patterns, which poses a significant challenge for accurate relocalization. In this scenario, our single-mode and sequence-mode approaches improve the recall rate of ACE by 7.2\% and 13.6\%, respectively. This highlights the effectiveness of the sequence-based mapping and relocalization methods, contributing recall rate improvements of 8.9\% and 4.8\%, respectively. \fref{fig:stairs} illustrates scene images and the cumulative error distribution from 0 to 1cm for different systems under this challenging scenario.

\begin{table}[t]
    \vspace{0.em}
    \setlength\tabcolsep{3.5pt}
    \caption{Comparison on the 12-Scenes dataset. We report the percentage of test frames below a (1cm, 1$^\circ$) pose error.}
    \label{tab:12-scenes}
    \centering  
    \scriptsize 
    \begin{tabular}{C{0.2\linewidth}|C{0.11\linewidth}C{0.09\linewidth}C{0.1\linewidth}C{0.14\linewidth}|C{0.07\linewidth}C{0.07\linewidth}}
        \toprule
        \multirow{2}{*}{Sequence} & \multirow{2}{*}{DSAC*\cite{brachmann2021visual}} & \multirow{2}{*}{ACE\cite{brachmann2023accelerated}}  & Focus- & \multirow{2}{*}{GLACE\cite{wang2024glace}} &  \multicolumn{2}{c}{Ours} \\
        & & & Tune\cite{nguyen2024focustune}  & &  Single &  Seq. \\
        \midrule
        Apt1\_kitchen      & 87.1 & 84.9 & 88.0 & 93.3 & 95.2 & \textbf{97.5}  \\
        Apt1\_living       & 91.3 & 89.5 & 85.2 & 92.9 & 90.9 & \textbf{95.9} \\
        Apt2\_bed          & 89.3 & 83.6 & 88.5 & 88.5 & 88.9 & \textbf{90.2} \\
        Apt2\_kitchen      & 83.0 & 84.3 & 90.0 & \textbf{93.0} & 84.3 & 90.9 \\
        Apt2\_living       & 83.6 & 89.1 & 89.4 & 91.1 & \textbf{92.5} & 91.6 \\
        Apt2\_luke         & 66.5 & 76.4 & 77.1 & 82.7 & 82.1 & \textbf{84.8}  \\
        Office1\_gates362  & 73.1 & 73.6 & 76.2 & \textbf{94.6} & 78.0 & 81.1 \\
        Office1\_gates381  & 62.3 & 68.2 & 70.0 & \textbf{85.2} & 74.0 & 78.8  \\
        Office1\_lounge    & 75.8 & 70.6 & 69.1 & \textbf{82.3} & 72.2 & 81.7 \\
        Office1\_manolis   & 70.9 & 72.9 & 71.4 & \textbf{84.4} & 72.7 & 77.2 \\
        Office2\_5a        & 62.6 & 60.2 & 54.1 & 68.6 & 68.6 & \textbf{73.8} \\
        Office2\_5b        & 58.3 & 71.4 & 61.2 & 84.0 & 77.5 & \textbf{85.9} \\
        \rowcolor{gray!20}
        Average   & 75.3 & 77.1 & 76.7 & \textbf{86.7} & 81.4 & 85.7 \\
        \midrule
        FPS     & 48 & 53 & 56 & 33 & \textbf{87} & 60 \\
        \bottomrule
    \end{tabular}
    \vspace{-1.0em}
\end{table}

\begin{table}[t]
    \vspace{0.em}
    \setlength\tabcolsep{3.5pt}
    \caption{Comparison on the Cambridge Landmarks dataset. Metrics include the median pose error and the recall rate.}
    \label{tab:cambridge}
    \centering  
    \scriptsize 
    \begin{threeparttable}
    \begin{tabular}{C{0.18\linewidth}|C{0.09\linewidth}C{0.13\linewidth}C{0.09\linewidth}C{0.09\linewidth}|C{0.09\linewidth}C{0.09\linewidth}}
        \toprule
        \multirow{2}{*}{Sequence}  & \multicolumn{4}{c|}{Median Error (cm / $^\circ$)} & \multicolumn{2}{c}{Ensemble ($\times$4)}  \\
        & HLoc\cite{sarlin2019coarse} & GLACE\cite{wang2024glace} & ACE\cite{brachmann2023accelerated} & Ours\tnote{\dag} & ACE\cite{brachmann2023accelerated} & Ours\tnote{\dag} \\
        \midrule
        GreatCourt    & \textbf{16}/\textbf{0.1} & 19/0.1 & 43/0.2 & 38/0.2 & 28/0.1  & 27/0.1\\
        KingsCollege  & \textbf{12}/\textbf{0.2} & 19/0.3 & 28/0.4 & 25/0.4 & 18/0.3 & 17/0.3 \\
        OldHospital   & \textbf{15}/\textbf{0.3} & 17/0.4 & 31/0.6 & 27/0.6 & 25/0.5 & 24/0.5 \\
        ShopFacade    & \textbf{4}/\textbf{0.2}  & \textbf{4}/\textbf{0.2}  & 5/0.3  & 5/0.3  & 5/0.3 & 5/0.3 \\
        StMarysChurch & \textbf{7}/\textbf{0.2}  & 9/0.3  & 18/0.6 & 17/0.6 & 9/0.3 & 9/0.3 \\
        \midrule
        \rowcolor{gray!20}
        Average       & \textbf{11}/\textbf{0.2} & 14/0.3 & 25/0.4 & 22/0.4 & 17/0.3 & 16/0.3 \\
        \bottomrule
    \end{tabular}
        \begin{tablenotes}
            \footnotesize
            \item[\dag] In this table, ``Ours" represents ``Ours-Single."
        \end{tablenotes}
    \end{threeparttable}
    \vspace{0.5em}
\end{table}

\begin{figure*}[t]
    \vspace{0.5em}
    \centering
    \setlength{\abovecaptionskip}{0.2cm}
    \includegraphics[width=0.9\linewidth]{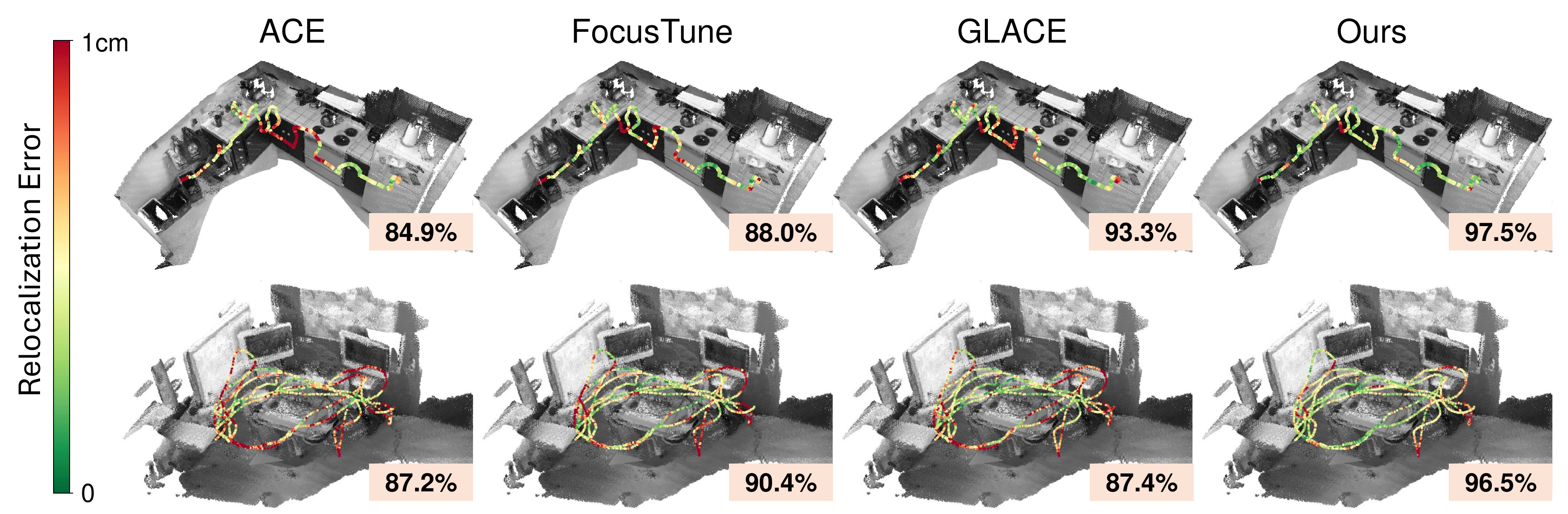}
    \caption{The relocalization trajectories of different SCR systems in two scenarios (top: Apt1\_kitchen, bottom: Chess). We also report the recall rates, defined as the percentage of test frames below a (1cm, 1\degree) pose error.
    }
    \label{fig:trajectory}
    \vspace{-2.3em}
\end{figure*}

\paragraph{12-Scenes}
The 12-Scenes dataset provides $1296 \times 968$ resolution image sequences across 12 scenes. Following prior works, we resize images to $642 \times 480$ for relocalization. \tref{tab:12-scenes} presents the evaluation results against SOTA SCR methods.
Compared to the baseline (ACE), our methods improve relocalization recall by 8.6\% and boost relocalization speed by 7$\hertz$. Although GLACE achieves slightly higher accuracy on this dataset, it comes at a significant cost: requiring 2.5$\times$ the mapping time, consuming 3$\times$ the memory for map storage, and operating at only half the relocalization speed of our approach. In \fref{fig:trajectory}, we compare the trajectory errors and recall rates of different algorithms for relocalization in two scenes.

\paragraph{Cambridge Landmarks}
This dataset comprises five scenes from the old town of Cambridge. The captured videos were downsampled into low-frame-rate image sequences, leading to large motion gaps between consecutive frames. These gaps make feature tracking via optical flow challenging, so we evaluate only our single-mode relocalization on this dataset. 
The results, summarized in \tref{tab:cambridge}, show that Ours-Single reduces the average error of ACE by 3cm, 
demonstrating the effectiveness of leveraging sequential information in enhancing mapping accuracy.
However, when using the ensemble method, the improvement of Ours-Single over ACE becomes minimal. This may be because the clustering process disrupts the original image sequence order, resulting in very limited sequence information available for Ours-Single.

\begin{table}[t]
    \vspace{0.em}
    \setlength\tabcolsep{3.5pt}
    \caption{Generalization validation.}
    \label{tab:seq_glace}
    \centering  
    \scriptsize 
    \begin{threeparttable}
    \begin{tabular}{C{0.25\linewidth}|C{0.15\linewidth}C{0.15\linewidth}C{0.2\linewidth}}
        \toprule
         \multirow{2}{*}{Methods} & 7-Scenes & 12-Scenes & Cambridge \\
         & (\%, \textcolor{ForestGreen}{$\uparrow$}) & (\%, \textcolor{ForestGreen}{$\uparrow$}) & (cm / $^\circ$, \textcolor{ForestGreen}{$\downarrow$}) \\
        \midrule
        ACE          &   55.2  & 77.1 & 25 / \textbf{0.4}  \\
        ACE + Ours   &   \textbf{66.2}  & \textbf{85.7} & \textbf{22} / \textbf{0.4}  \\
        \midrule
        GLACE          &  61.6  & 86.7 & 14 / \textbf{0.3}   \\
        GLACE + Ours   &  \textbf{69.7}  & \textbf{89.8} & \textbf{13} / \textbf{0.3}   \\
        \bottomrule
    \end{tabular}
        \begin{tablenotes}
            \footnotesize
            \item[1] We report the percentage of test frames below a (1cm, 1$^\circ$) pose error on 7-Scenes and 12-Scenes, as well as the median pose errors on Cambridge Landmarks.
        \end{tablenotes}
    \end{threeparttable}
    \vspace{0.3em}
\end{table}

\subsection{Generalization Validation}\label{sec:seq_test}
Although our system builds upon ACE \cite{brachmann2023accelerated}, the proposed methods are not dependent on the specific structure. To validate this, we also apply our methods to another SCR system, GLACE \cite{wang2024glace}.
As shown in \tref{tab:seq_glace}, our methods yield improvements on both ACE and GLACE, where "ACE + Ours" and "GLACE + Ours" denote the application of our methods to ACE and GLACE, respectively. The results indicate consistent performance gains across all three datasets. However, the improvement on Cambridge Landmarks is relatively smaller, likely due to its limited sequential information compared to the 7-Scenes and 12-Scenes datasets, as discussed in \sref{sec:accuracy}.
Overall, these experimental results highlight both the effectiveness and generalization capability of the proposed methods.


\begin{table}[t]
    \vspace{0.em}
    \setlength\tabcolsep{3.5pt}
    \caption{Ablation study on the proposed keypoint detection head, sequence-based mapping, and sequence-based relocalization.}
    \label{tab:ablation}
    \centering  
    \scriptsize 
    \arrayrulecolor{black}  
    \color{black}           
    \begin{threeparttable}
    \begin{tabular}{C{0.3\linewidth}|C{0.15\linewidth}C{0.08\linewidth}|C{0.15\linewidth}C{0.08\linewidth}}
        \toprule
        \multirow{2}{*}{Methods\tnote{\dag}}  & \multicolumn{2}{c|}{7-Scenes} & \multicolumn{2}{c}{12-Scenes}  \\
        &  Recall (\%) & \textcolor{ForestGreen}{$\Uparrow$} &  Recall (\%) & \textcolor{ForestGreen}{$\Uparrow$} \\
        \midrule
        w/o KDH, SM, SR     & 55.2 &  & 77.1  & \\
        w/o SM, SR         & 60.3 & \textcolor{ForestGreen}{+5.1} & 80.4  & \textcolor{ForestGreen}{+3.3}\\
        w/o SR             & 61.6 & \textcolor{ForestGreen}{+6.4} & 81.4  & \textcolor{ForestGreen}{+4.3}\\
        \midrule
        Ours (Full)              & \textbf{66.2} & \textcolor{ForestGreen}{+11.0} & \textbf{85.7} & \textcolor{ForestGreen}{+8.6} \\
        \bottomrule
    \end{tabular}
        \begin{tablenotes}
            \footnotesize
            \item[\dag] KDH, SM, and SR represent keypoint detection head, sequence-based mapping, and sequence-based relocalization, respectively.
        \end{tablenotes}
    \end{threeparttable}
    \vspace{0.3em}
\end{table} 

\subsection{Ablation Study}\label{sec:ablation}


We perform an ablation study on the 7-Scenes and 12-Scenes datasets to assess the contributions of the proposed keypoint detection head (KDH), sequence-based mapping (SM), and sequence-based relocalization (SR). By sequentially removing these modules, we compute the recall rate, defined as the percentage of test frames with a pose error below (1cm, 1\degree). The results in \tref{tab:ablation} demonstrate the effectiveness of the proposed components. Since our system builds on ACE, the ``w/o KDH, SM, SR'' setting corresponds to ACE. On the 7-Scenes dataset, KDH, SM, and SR improve recall by 5.1\%, 1.3\%, and 4.6\%, respectively, achieving a total improvement of 11.0\%. On the 12-Scenes dataset, these methods yield an 8.6\% gain, further validating their impact.

\section{Conclusion}

In this work, we propose an efficient and accurate SCR system with a unified scene-agnostic module for scene encoding and keypoint detection, prioritizing informative regions. To enhance implicit triangulation, we associate keypoints across images to leverage sequential information. Additionally, we introduce two relocalization modes: a high-efficiency single-mode and a sequence-mode that integrates temporal cues for improved accuracy. Experimental results demonstrate that our methods significantly improve system performance, achieving remarkable accuracy and efficiency.
Nevertheless, like other SCR systems, we use a CNN as the backbone, which limits scene point predictions to rely solely on information from a single local patch. In contrast, ViT \cite{dosovitskiy2020vit} can capture long-range dependencies between image patches and leverage image global information to some extent. 
Therefore, for future work, we suggest training a scene-agnostic ViT encoder to extract feature vectors, enhancing the SCR system’s ability to leverage global image information.



\bibliographystyle{IEEEtran}
\bibliography{IEEEabrv}

\begin{thebibliography}{10}
\providecommand{\url}[1]{#1}
\csname url@rmstyle\endcsname
\providecommand{\newblock}{\relax}
\providecommand{\bibinfo}[2]{#2}
\providecommand\BIBentrySTDinterwordspacing{\spaceskip=0pt\relax}
\providecommand\BIBentryALTinterwordstretchfactor{4}
\providecommand\BIBentryALTinterwordspacing{\spaceskip=\fontdimen2\font plus
\BIBentryALTinterwordstretchfactor\fontdimen3\font minus \fontdimen4\font\relax}
\providecommand\BIBforeignlanguage[2]{{%
\expandafter\ifx\csname l@#1\endcsname\relax
\typeout{** WARNING: IEEEtran.bst: No hyphenation pattern has been}%
\typeout{** loaded for the language `#1'. Using the pattern for}%
\typeout{** the default language instead.}%
\else
\language=\csname l@#1\endcsname
\fi
#2}}

\bibitem{brachmann2021visual}
E.~Brachmann and C.~Rother, ``Visual camera re-localization from rgb and rgb-d images using dsac,'' \emph{IEEE transactions on pattern analysis and machine intelligence}, vol.~44, no.~9, pp. 5847--5865, 2021.

\bibitem{schneider2018maplab}
T.~Schneider, M.~Dymczyk, M.~Fehr, K.~Egger, S.~Lynen, I.~Gilitschenski, and R.~Siegwart, ``maplab: An open framework for research in visual-inertial mapping and localization,'' \emph{IEEE Robotics and Automation Letters}, vol.~3, no.~3, pp. 1418--1425, 2018.

\bibitem{sarlin2019coarse}
P.-E. Sarlin, C.~Cadena, R.~Siegwart, and M.~Dymczyk, ``From coarse to fine: Robust hierarchical localization at large scale,'' in \emph{CVPR}, 2019.

\bibitem{kendall2017geometric}
A.~Kendall and R.~Cipolla, ``Geometric loss functions for camera pose regression with deep learning,'' in \emph{Proceedings of the IEEE conference on computer vision and pattern recognition}, 2017, pp. 5974--5983.

\bibitem{schoenberger2016sfm}
J.~L. Sch\"{o}nberger and J.-M. Frahm, ``Structure-from-motion revisited,'' in \emph{Conference on Computer Vision and Pattern Recognition (CVPR)}, 2016.

\bibitem{xu2025airslam}
K.~Xu, Y.~Hao, S.~Yuan, C.~Wang, and L.~Xie, ``Airslam: An efficient and illumination-robust point-line visual slam system,'' \emph{IEEE Transactions on Robotics}, 2025.

\bibitem{dong2022visual}
S.~Dong, S.~Wang, Y.~Zhuang, J.~Kannala, M.~Pollefeys, and B.~Chen, ``Visual localization via few-shot scene region classification,'' in \emph{2022 International Conference on 3D Vision (3DV)}.\hskip 1em plus 0.5em minus 0.4em\relax IEEE, 2022, pp. 393--402.

\bibitem{brachmann2023accelerated}
E.~Brachmann, T.~Cavallari, and V.~A. Prisacariu, ``Accelerated coordinate encoding: Learning to relocalize in minutes using rgb and poses,'' in \emph{Proceedings of the IEEE/CVF Conference on Computer Vision and Pattern Recognition}, 2023, pp. 5044--5053.

\bibitem{nguyen2024focustune}
S.~T. Nguyen, A.~Fontan, M.~Milford, and T.~Fischer, ``Focustune: Tuning visual localization through focus-guided sampling,'' in \emph{Proceedings of the IEEE/CVF Winter Conference on Applications of Computer Vision}, 2024, pp. 3606--3615.

\bibitem{wang2024glace}
F.~Wang, X.~Jiang, S.~Galliani, C.~Vogel, and M.~Pollefeys, ``Glace: Global local accelerated coordinate encoding,'' in \emph{Proceedings of the IEEE/CVF Conference on Computer Vision and Pattern Recognition}, 2024, pp. 21\,562--21\,571.

\bibitem{zhang2023map}
Z.~Zhang, Y.~Jiao, S.~Huang, R.~Xiong, and Y.~Wang, ``Map-based visual-inertial localization: Consistency and complexity,'' \emph{IEEE Robotics and Automation Letters}, vol.~8, no.~3, pp. 1407--1414, 2023.

\bibitem{liu2024pe}
C.~Liu, H.~Yu, P.~Cheng, W.~Sun, J.~Civera, and X.~Chen, ``Pe-vins: Accurate monocular visual-inertial slam with point-edge features,'' \emph{IEEE Transactions on Intelligent Vehicles}, 2024.

\bibitem{zhou2022geometry}
Q.~Zhou, S.~Agostinho, A.~O{\v{s}}ep, and L.~Leal-Taix{\'e}, ``Is geometry enough for matching in visual localization?'' in \emph{European Conference on Computer Vision}.\hskip 1em plus 0.5em minus 0.4em\relax Springer, 2022, pp. 407--425.

\bibitem{sarlin2021back}
P.-E. Sarlin, A.~Unagar, M.~Larsson, H.~Germain, C.~Toft, V.~Larsson, M.~Pollefeys, V.~Lepetit, L.~Hammarstrand, F.~Kahl, \emph{et~al.}, ``Back to the feature: Learning robust camera localization from pixels to pose,'' in \emph{Proceedings of the IEEE/CVF conference on computer vision and pattern recognition}, 2021, pp. 3247--3257.

\bibitem{kendall2015posenet}
A.~Kendall, M.~Grimes, and R.~Cipolla, ``Posenet: A convolutional network for real-time 6-dof camera relocalization,'' in \emph{Proceedings of the IEEE international conference on computer vision}, 2015, pp. 2938--2946.

\bibitem{melekhov2017image}
I.~Melekhov, J.~Ylioinas, J.~Kannala, and E.~Rahtu, ``Image-based localization using hourglass networks,'' in \emph{Proceedings of the IEEE international conference on computer vision workshops}, 2017, pp. 879--886.

\bibitem{clark2017vidloc}
R.~Clark, S.~Wang, A.~Markham, N.~Trigoni, and H.~Wen, ``Vidloc: A deep spatio-temporal model for 6-dof video-clip relocalization,'' in \emph{Proceedings of the IEEE conference on computer vision and pattern recognition}, 2017, pp. 6856--6864.

\bibitem{purkait2018synthetic}
P.~Purkait, C.~Zhao, and C.~Zach, ``Synthetic view generation for absolute pose regression and image synthesis.'' in \emph{BMVC}, 2018, p.~69.

\bibitem{chen2021direct}
S.~Chen, Z.~Wang, and V.~Prisacariu, ``Direct-posenet: Absolute pose regression with photometric consistency,'' in \emph{2021 International Conference on 3D Vision (3DV)}.\hskip 1em plus 0.5em minus 0.4em\relax IEEE, 2021, pp. 1175--1185.

\bibitem{cai2019camera}
M.~Cai, H.~Zhan, C.~Saroj~Weerasekera, K.~Li, and I.~Reid, ``Camera relocalization by exploiting multi-view constraints for scene coordinates regression,'' in \emph{Proceedings of the IEEE/CVF International Conference on Computer Vision Workshops}, 2019, pp. 0--0.

\bibitem{tang2021learning}
S.~Tang, C.~Tang, R.~Huang, S.~Zhu, and P.~Tan, ``Learning camera localization via dense scene matching,'' in \emph{Proceedings of the IEEE/CVF Conference on Computer Vision and Pattern Recognition}, 2021, pp. 1831--1841.

\bibitem{zhou2020kfnet}
L.~Zhou, Z.~Luo, T.~Shen, J.~Zhang, M.~Zhen, Y.~Yao, T.~Fang, and L.~Quan, ``Kfnet: Learning temporal camera relocalization using kalman filtering,'' in \emph{Proceedings of the IEEE/CVF conference on computer vision and pattern recognition}, 2020, pp. 4919--4928.

\bibitem{kalman1960new}
R.~E. Kalman, ``A new approach to linear filtering and prediction problems,'' 1960.

\bibitem{bui2024d2s}
B.-T. Bui, H.-H. Bui, D.-T. Tran, and J.-H. Lee, ``D2s: Representing sparse descriptors and 3d coordinates for camera relocalization,'' \emph{IEEE Robotics and Automation Letters}, 2024.

\bibitem{brachmann2018learning}
E.~Brachmann and C.~Rother, ``Learning less is more-6d camera localization via 3d surface regression,'' in \emph{Proceedings of the IEEE conference on computer vision and pattern recognition}, 2018, pp. 4654--4662.

\bibitem{liu2024reprojection}
T.-R. Liu, H.-K. Yang, J.-M. Liu, C.-W. Huang, T.-C. Chiang, Q.~Kong, N.~Kobori, and C.-Y. Lee, ``Reprojection errors as prompts for efficient scene coordinate regression,'' in \emph{European Conference on Computer Vision}.\hskip 1em plus 0.5em minus 0.4em\relax Springer, 2024, pp. 286--302.

\bibitem{viswanathan2009features}
D.~G. Viswanathan, ``Features from accelerated segment test (fast),'' in \emph{Proceedings of the 10th workshop on image analysis for multimedia interactive services, London, UK}, 2009, pp. 6--8.

\bibitem{rublee2011orb}
E.~Rublee, V.~Rabaud, K.~Konolige, and G.~Bradski, ``{ORB}: An efficient alternative to sift or surf,'' in \emph{2011 International conference on computer vision}.\hskip 1em plus 0.5em minus 0.4em\relax IEEE, 2011, pp. 2564--2571.

\bibitem{detone2018superpoint}
D.~DeTone, T.~Malisiewicz, and A.~Rabinovich, ``Superpoint: Self-supervised interest point detection and description,'' in \emph{Proceedings of the IEEE conference on computer vision and pattern recognition workshops}, 2018, pp. 224--236.

\bibitem{revaud2019r2d2}
J.~Revaud, C.~De~Souza, M.~Humenberger, and P.~Weinzaepfel, ``R2d2: Reliable and repeatable detector and descriptor,'' \emph{Advances in neural information processing systems}, vol.~32, 2019.

\bibitem{dusmanu2019d2}
M.~Dusmanu, I.~Rocco, T.~Pajdla, M.~Pollefeys, J.~Sivic, A.~Torii, and T.~Sattler, ``D2-net: A trainable cnn for joint description and detection of local features,'' in \emph{Proceedings of the ieee/cvf conference on computer vision and pattern recognition}, 2019, pp. 8092--8101.

\bibitem{dai2017scannet}
A.~Dai, A.~X. Chang, M.~Savva, M.~Halber, T.~Funkhouser, and M.~Nie{\ss}ner, ``Scannet: Richly-annotated 3d reconstructions of indoor scenes,'' in \emph{Proc. Computer Vision and Pattern Recognition (CVPR), IEEE}, 2017.

\bibitem{chen2018gradnorm}
Z.~Chen, V.~Badrinarayanan, C.-Y. Lee, and A.~Rabinovich, ``Gradnorm: Gradient normalization for adaptive loss balancing in deep multitask networks,'' in \emph{International conference on machine learning}.\hskip 1em plus 0.5em minus 0.4em\relax PMLR, 2018, pp. 794--803.

\bibitem{hartley2003multiple}
R.~Hartley and A.~Zisserman, \emph{Multiple view geometry in computer vision}.\hskip 1em plus 0.5em minus 0.4em\relax Cambridge university press, 2003.

\bibitem{lindenberger2023lightglue}
P.~Lindenberger, P.-E. Sarlin, and M.~Pollefeys, ``Lightglue: Local feature matching at light speed,'' in \emph{Proceedings of the IEEE/CVF International Conference on Computer Vision}, 2023, pp. 17\,627--17\,638.

\bibitem{loshchilov2018decoupled}
I.~Loshchilov and F.~Hutter, ``Decoupled weight decay regularization,'' in \emph{International Conference on Learning Representations}, 2019.

\bibitem{smith2019super}
L.~N. Smith and N.~Topin, ``Super-convergence: Very fast training of neural networks using large learning rates,'' in \emph{Artificial intelligence and machine learning for multi-domain operations applications}, vol. 11006.\hskip 1em plus 0.5em minus 0.4em\relax SPIE, 2019, pp. 369--386.

\bibitem{xu2023airvo}
K.~Xu, Y.~Hao, S.~Yuan, C.~Wang, and L.~Xie, ``Airvo: An illumination-robust point-line visual odometry,'' in \emph{2023 IEEE/RSJ International Conference on Intelligent Robots and Systems (IROS)}.\hskip 1em plus 0.5em minus 0.4em\relax IEEE, 2023, pp. 3429--3436.

\bibitem{lucas1981iterative}
B.~D. Lucas and T.~Kanade, ``An iterative image registration technique with an application to stereo vision,'' in \emph{IJCAI'81: 7th international joint conference on Artificial intelligence}, vol.~2, 1981, pp. 674--679.

\bibitem{shotton2013scene}
J.~Shotton, B.~Glocker, C.~Zach, S.~Izadi, A.~Criminisi, and A.~Fitzgibbon, ``Scene coordinate regression forests for camera relocalization in rgb-d images,'' in \emph{Proceedings of the IEEE conference on computer vision and pattern recognition}, 2013, pp. 2930--2937.

\bibitem{valentin2016learning}
J.~Valentin, A.~Dai, M.~Nie{\ss}ner, P.~Kohli, P.~Torr, S.~Izadi, and C.~Keskin, ``Learning to navigate the energy landscape,'' in \emph{2016 Fourth International Conference on 3D Vision (3DV)}.\hskip 1em plus 0.5em minus 0.4em\relax IEEE, 2016, pp. 323--332.

\bibitem{dosovitskiy2020vit}
A.~Dosovitskiy, L.~Beyer, A.~Kolesnikov, D.~Weissenborn, X.~Zhai, T.~Unterthiner, M.~Dehghani, M.~Minderer, G.~Heigold, S.~Gelly, J.~Uszkoreit, and N.~Houlsby, ``An image is worth 16x16 words: Transformers for image recognition at scale,'' \emph{ICLR}, 2021.

\end{thebibliography}


\end{document}